\documentclass{article}

\usepackage{arxiv}

\usepackage[utf8]{inputenc} 
\usepackage[T1]{fontenc}    
\usepackage{hyperref}       
\usepackage{url}            
\usepackage{booktabs}       
\usepackage{amsfonts}       
\usepackage{nicefrac}       
\usepackage{microtype}      
\usepackage{lipsum}		
\usepackage{graphicx}
\usepackage{doi}
\usepackage{amsmath} 
\usepackage{amssymb}  
\usepackage{subfigure}
\usepackage{parskip}

\title{Real-World Dexterous Object Manipulation based Deep Reinforcement Learning}

\author{Westlake University\\
Qingfeng Yao, Jilong Wang, Shuyu Yang\\
}

\date{}

\begin{document}
\maketitle

\begin{abstract}
Deep reinforcement learning has shown its advantages in real-time decision-making based on the state of the agent. 
In this stage, we solved the task of using a real robot to manipulate the cube to a given trajectory. The task is broken down into different procedures and we propose a hierarchical structure, the high-level deep reinforcement learning model selects appropriate contact positions and the low-level control module performs the position control under the corresponding trajectory. Our framework reduces the disadvantage of low sample efficiency of deep reinforcement learning and lacking adaptability of traditional robot control methods.
Our algorithm is trained in simulation and migrated to reality without fine-tuning. The experimental results show the effectiveness of our method both simulation and reality.
Our code and video can be found at https://github.com/42jaylonw/RRC2021ThreeWolves and https://youtu.be/Jr176xsn9wg.
\end{abstract}

\section{Introduction}
The Real Robot Challenge is a manipulation challenge using the TriFinger Platform\cite{wuthrich2020trifinger}.
The TriFinger robot consists of three identical fingers, each with three degrees of freedom located above the robot's workspace. Our task is to pick up a cube and make it move along a given trajectory. We first manipulated the cube in a simulated environment. And furthermore test our method using a real platform.
\begin{figure}[ht]
\centering
\includegraphics[width=0.6\textwidth]{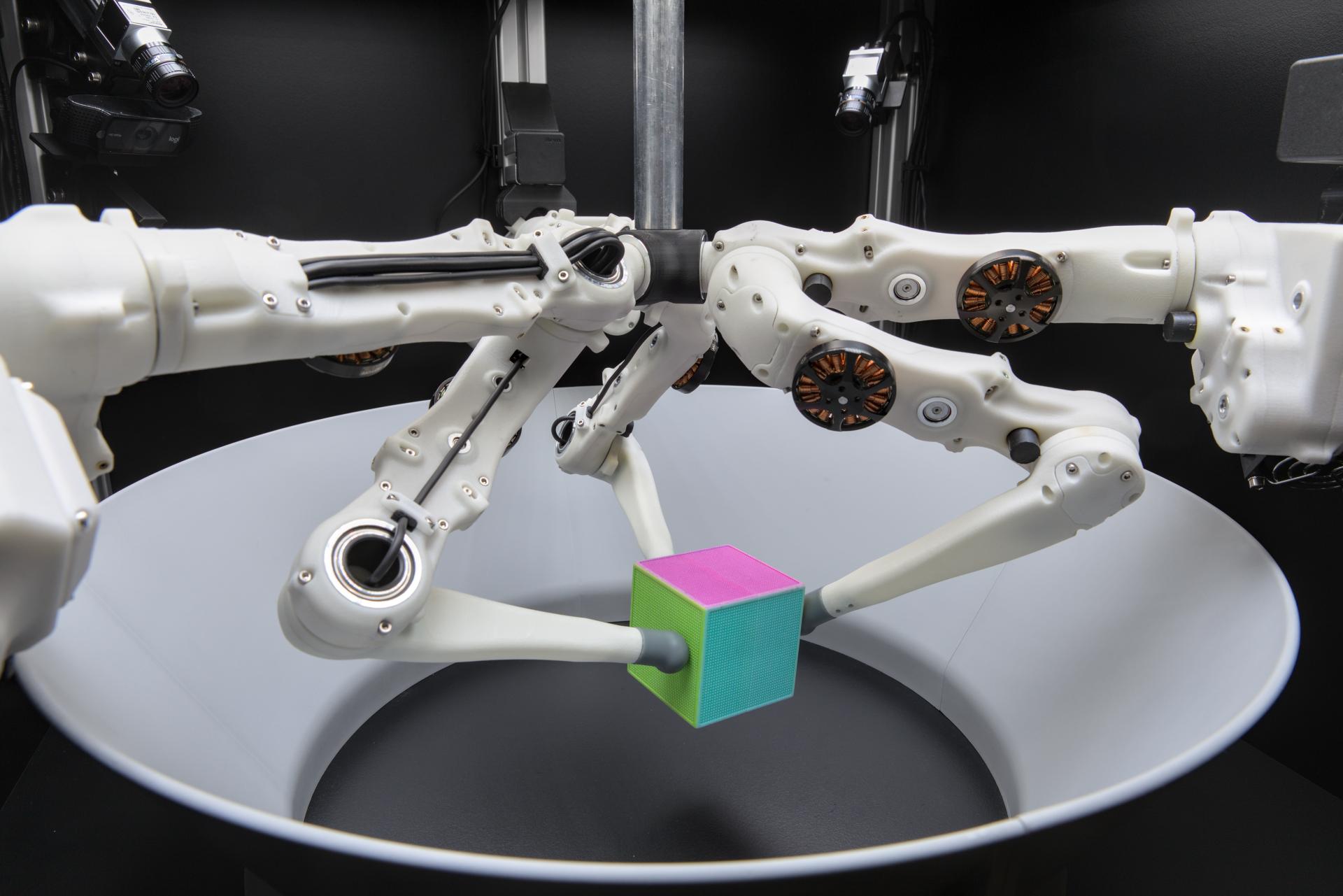}
\caption{The overall framework of our approach}
\label{fig:task}
\end{figure}

Our method relies on two points: 

(1) Manually set contact points is impossible to following different goal trajectory while grasping object stably. Such, we introduce the reinforcement learning to generate appreciate contact points by observing the object state and next goal position.



(2) Classic robot control method performs stably and accurately in the task of joint position control and trajectory following.

Therefore, we use a hierarchical control framework that utilize reinforcement learning as high-level planner and classic control methods as low-level controller. We broke down the task of grasping and moving the cube in given trajectory into three primitives: selecting three suitable contact points, moving the tip to the contact points of the cube, and finally lifting the cube to the given trajectory.

\section{Method}
\begin{figure}[ht]
\centering
\includegraphics[width=0.8\textwidth]{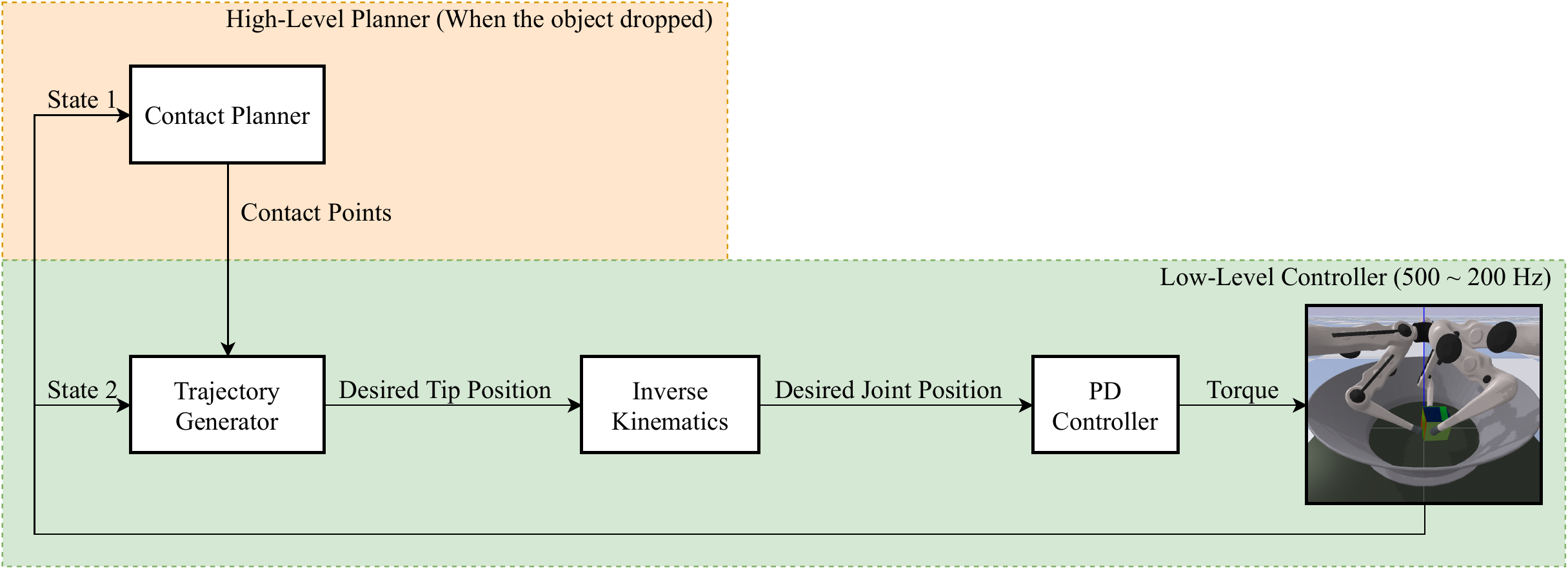}
\caption{The overall framework of our approach}
\label{fig:overall}
\end{figure}
\subsection{Deep Reinforcement Learning} 
DRL can be understood as a Markov Decision Process (MDP) denoted by a 5-tuple,
$\{\mathcal{S},\mathcal{A},\mathcal{P},\mathcal{R},\gamma \}$, where $\mathcal{S}$ is the state space, $\mathcal{A}$ is the action space, $\mathcal{P}$ is the state transition probability, $\mathcal{R}$ is the reward space, and $\gamma \in (0,1)$ is the discount factor. The robot starts with the state $s_0 \in \mathcal{S}$ with the probability of $p(s_0)$.
 At each time step $t$, the robot chooses an action $a_t\in\mathcal{A}$ with a probability of $\pi(a_t|s_t)$, and this action will be rewarded with $r(a_t,s_t)\in\mathcal{R}$. Then the robot will enter another state,
$s_{t+1}\in\mathcal{S}$, in the next time step with the probability of $\mathcal{P}=p(s_{t+1}|s_t,a_t)$. By repeating this process, we can generate
a trajectory of states, actions, and rewards $\tau=\{s_0,a_0,r_0,s_1,a_1,r_1,\cdots,s_t,a_t,r_t,\cdots\}$. On top of these, the Q-value, $Q(s, a)$, is defined to measure the quality of a state-action pair, usually denoted by the discounted cumulative sum of rewards along the trajectory.  In the DRL framework of SAC, the policy $\pi(a_t|s_t)$ and the Q-values $Q(s, a)$ are parameterized as neural networks, which are also known as the actor network and the critic network. 

\subsection{Entropy-Regularized DRL}\label{Entropy-Regularized DRL}
Entropy is a quantity that represents the randomness of a random variable, and it takes the form of:
\begin{equation}
H(P)=\underset{x \sim P}{\mathrm{E}}[-\log P(x)]
\end{equation}
As shown, $x$ is the value of a random variable with a probabilistic mass or density function, $P$, and the entropy $H$ of $x$ is calculated according to its distribution $P$.

In entropy-regularized DRL \cite{haarnoja2018soft}, agents obtain extra rewards proportional to the strategy entropy at each time step. Thus the optimized policy can be written as:

\begin{equation}\label{optimized pi}
\begin{aligned}
\pi^{*}=& \arg \max _{\pi} \underset{\tau \sim \pi}{\mathrm{E}}\left[\sum_{t=0}^{\infty} \gamma^{t}\left(R\left(s_{t}, a_{t}, s_{t+1}\right)\right.\right.\\
&\left.+\alpha H\left(\pi\left(\cdot \mid s_{t}\right)\right) \mid s_{0}=s\right]
\end{aligned}
\end{equation}
$\alpha$ is the trade-off coefficient to scale the effect of the information entropy term. The entropy term in (\ref{optimized pi}) explicitly encourages the robot to add uncertainty to its action strategy.
\subsection{DRL-based Contact Point Planner}
\begin{figure}[tb]
    \centering
    \subfigure[2D diagram]{
        \includegraphics[height=0.6in]{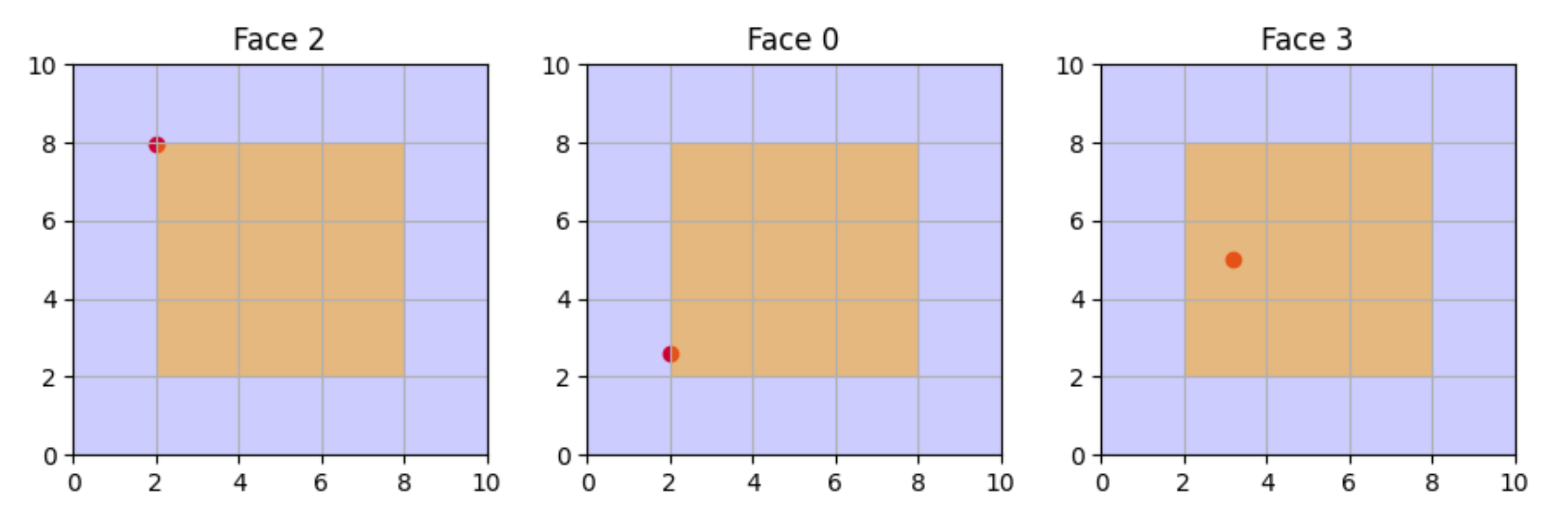}
    }
    \subfigure[3D sketch]{
	\includegraphics[height=0.6in]{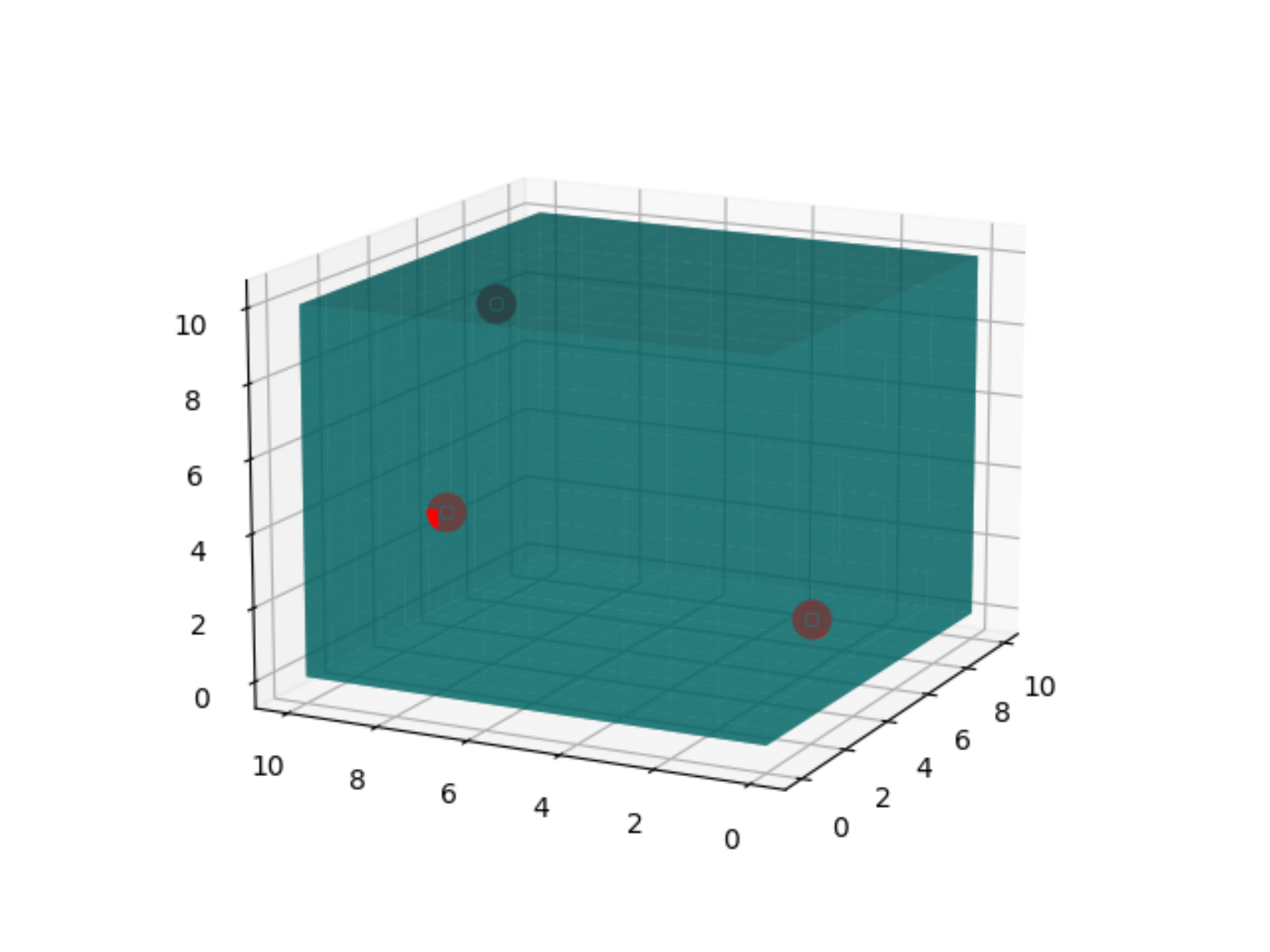}
    }
    \caption{2D and 3D sketch of the selected contact position, the orange area in the first diagram means the available contact area to select}
    \label{fig:3d}
\end{figure}
In this report, we complete the task with a reinforcement learning method and PD position controller.
Our framework is shown in Fig .\ref{fig:overall}. Reinforcement learning selects appropriate contact points, and the control algorithm completes the movement according to the generated trajectory. 
Considering the safe contact points, we set the available contact area in 60\% of one cube face from the center, as shown in Fig. \ref{fig:3d}, and corresponding expected contact positions are generated by the reinforcement learning and executed by a low-level PD controller.

In terms of multiple goal poses reaching the task, it’s important to select a controlling strategy based on different goals. It is critical to select the action of a tri-finger robot at grasping an object since its large and continuous working space. A successful strategy should give the action based on various goals, especially in the period of contact point choosing. Hardcoding contact points strategy often costs many trials of falling and re-grasping to reach on goal. To overcome this, we method use DRL that searches three optimal contact points on the object based on current robot states and current goal poses. Our framework is shown in Fig .1. Reinforcement learning selects appropriate contact points, and the control algorithm completes the movement according to the generated trajectory. Considering the safe contact points, we set the available contact area in 60\% of one cube face from the center, as shown in Fig. 2, and corresponding expected contact positions are generated by the reinforcement learning and executed by a low-level PD controller. 
The framework that we use is SAC\cite{haarnoja2018soft} . In the environment of moving the cube on a given trajectory, the robot needs to move the target object to the desired position fast and stably. When a reasonable contact point is selected, the subsequent control algorithm is more likely to reach the target point stably. On the contrary, if an unreasonable contact point is selected, it is difficult for the low-level controller algorithm to move the block to the target point, so we use the distance between the block and the trajectory as a reward to help the agent learn to choose the right contact point.

\subsection{Rigid Body Grasping Controller}
Tri-finger object grasping is profoundly similar to the quadruped locomotion task. Both of the configurations are formed by one flow body and few arms if you treat the object as flow body. Except there is disconnection when the tri-finger changes its contact points. Therefore we consider the goal pose reaching process as a flow rigid body with four supporting ‘legs’. We utilize quadratic programming to solve each force of tip for following the designed trajectory.



The framework that we use is SAC. 
In the environment of move the cube on a given trajectory, the robot need to move the target object to the desired position fast and stably. 
When a reasonable contact point is selected, the subsequent control algorithm is more likely to reach the target point stably. On the contrary, if an unreasonable contact point is selected, it is difficult for the low-level controller algorithm to move the block to the target point, so we use the distance between the block and the trajectory as a reward to help the agent learn to choose the right contact point.

\begin{equation}
r=0.001\exp(-300\left\|p_{goal}-p_{cube}\right\|^2)
\end{equation}
We generate the corresponding trajectory according to the current fingertip position and the target point. In order to keep the movement stable, we adopted fifth-order polynomials\cite{lynch2017modern}. 

We generate the corresponding trajectory according to the current fingertip position and the target point. In order to keep the movement stable, we adopted fifth-order polynomials. 
The robot is asked to avoid a discontinuous jump in acceleration at both $t = 0$ and $t = T$.
Our solution limits the terminal position, velocity, and acceleration, but adding these constraints to the problem formulation requires the addition of design freedoms in the polynomial, yielding a quintic polynomial of
time, $s(t)=a_{0}+\cdots+a_{5} t^{5}$. We can use the six terminal position, velocity, and acceleration constraints to solve uniquely for $s(0)=\dot{s}(0)=\ddot{s}(0)=\dot{s}(T)=\ddot{s}(T)=0$ and $s(T)=n$, which yields a smooth motion. 
In the final step, the PD controller and inverse dynamics are used to keep the corresponding fingertips to the desired trajectory.

\begin{figure}[ht]
\centering

\includegraphics[width=0.3\textwidth,height=0.1\textheight]{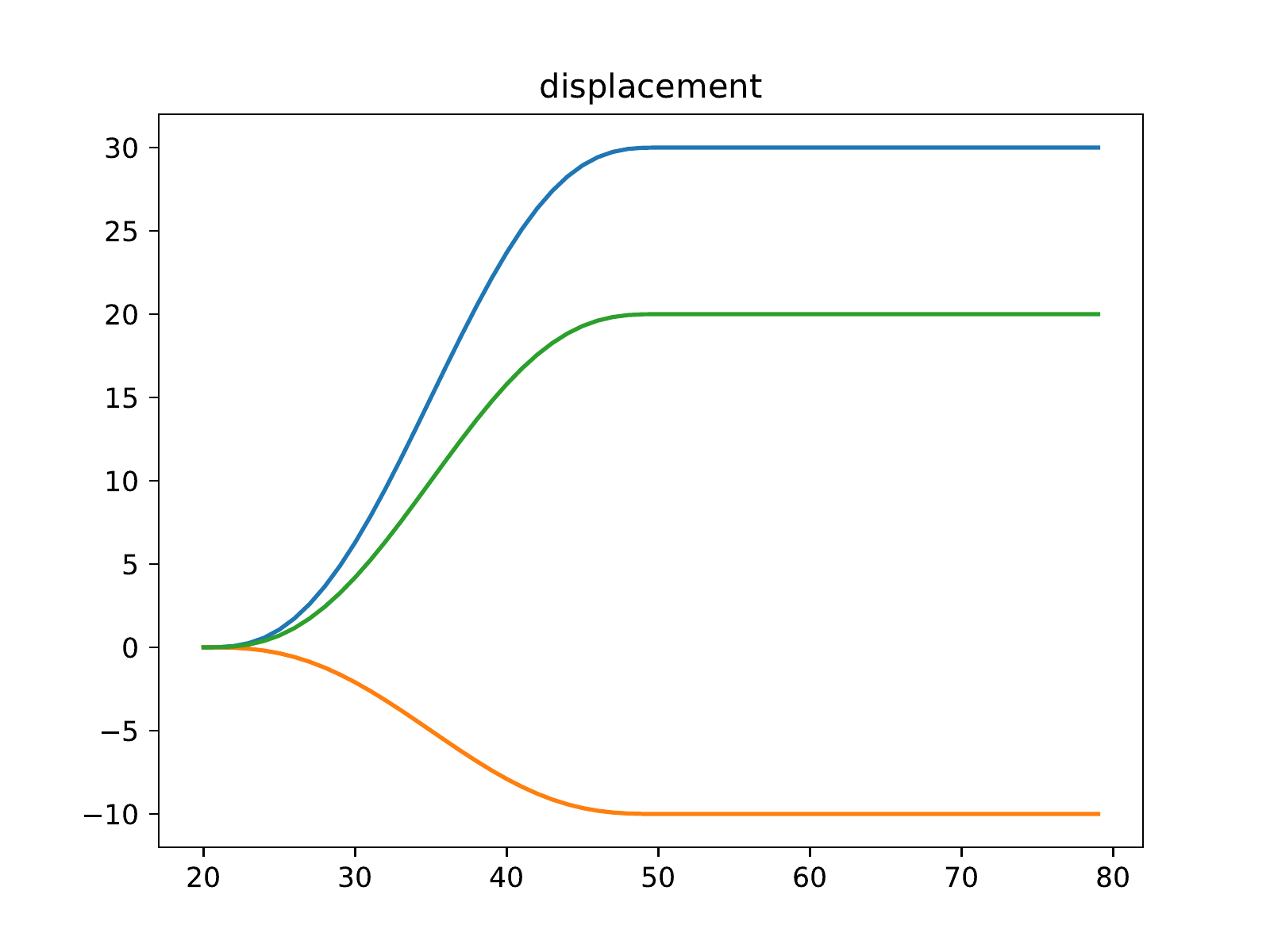}
\quad
\includegraphics[width=0.3\textwidth,height=0.1\textheight]{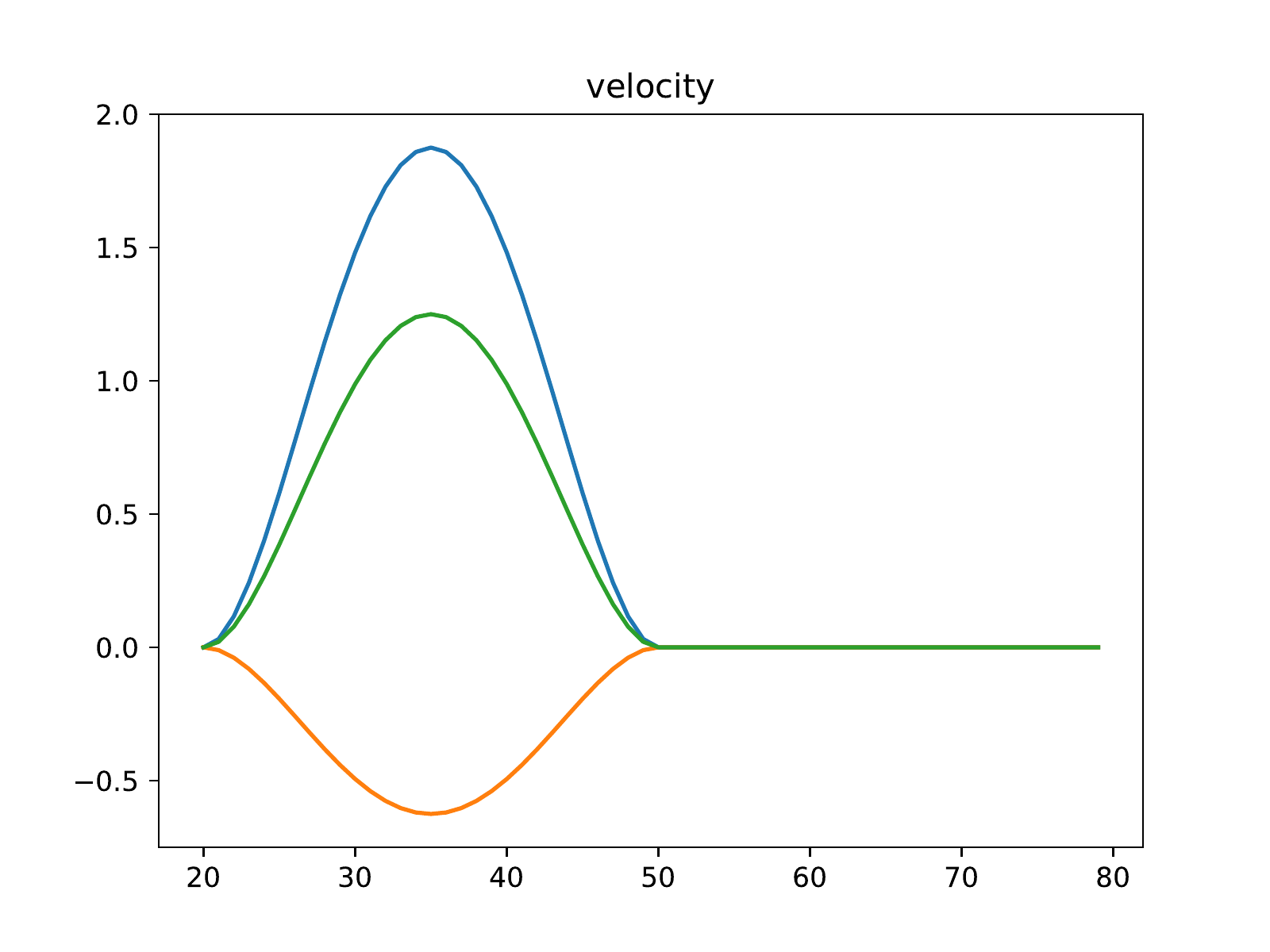}
\quad
\includegraphics[width=0.3\textwidth,height=0.1\textheight]{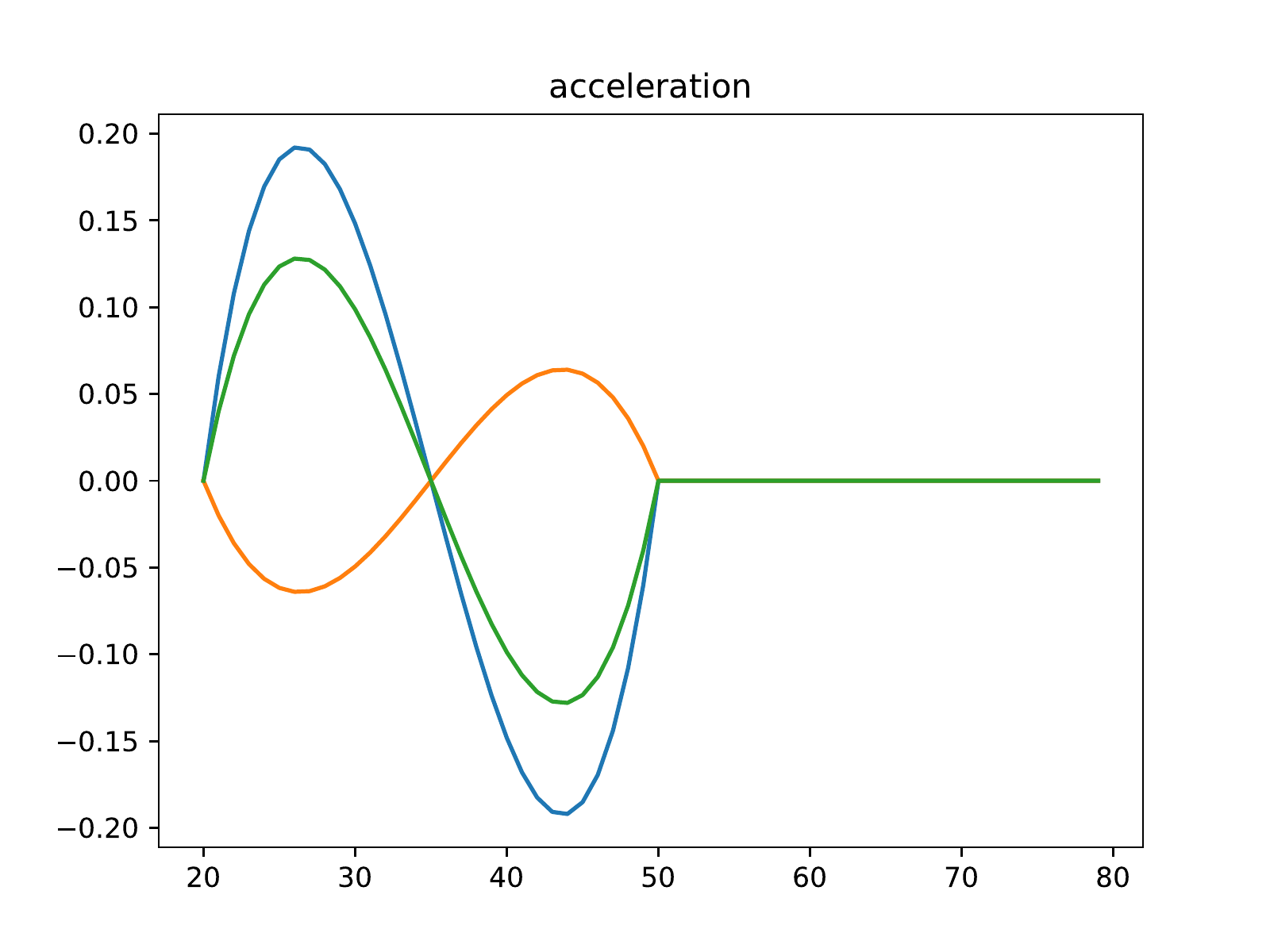}
\caption{
Trajectory, velocity, and acceleration curves based on fifth-order polynomials
}
\label{Fig.poly}
\end{figure}


\section{Result and Discussion}

\begin{figure}[ht]
\centering
\includegraphics[width=0.4\textwidth, height=0.2\textheight]{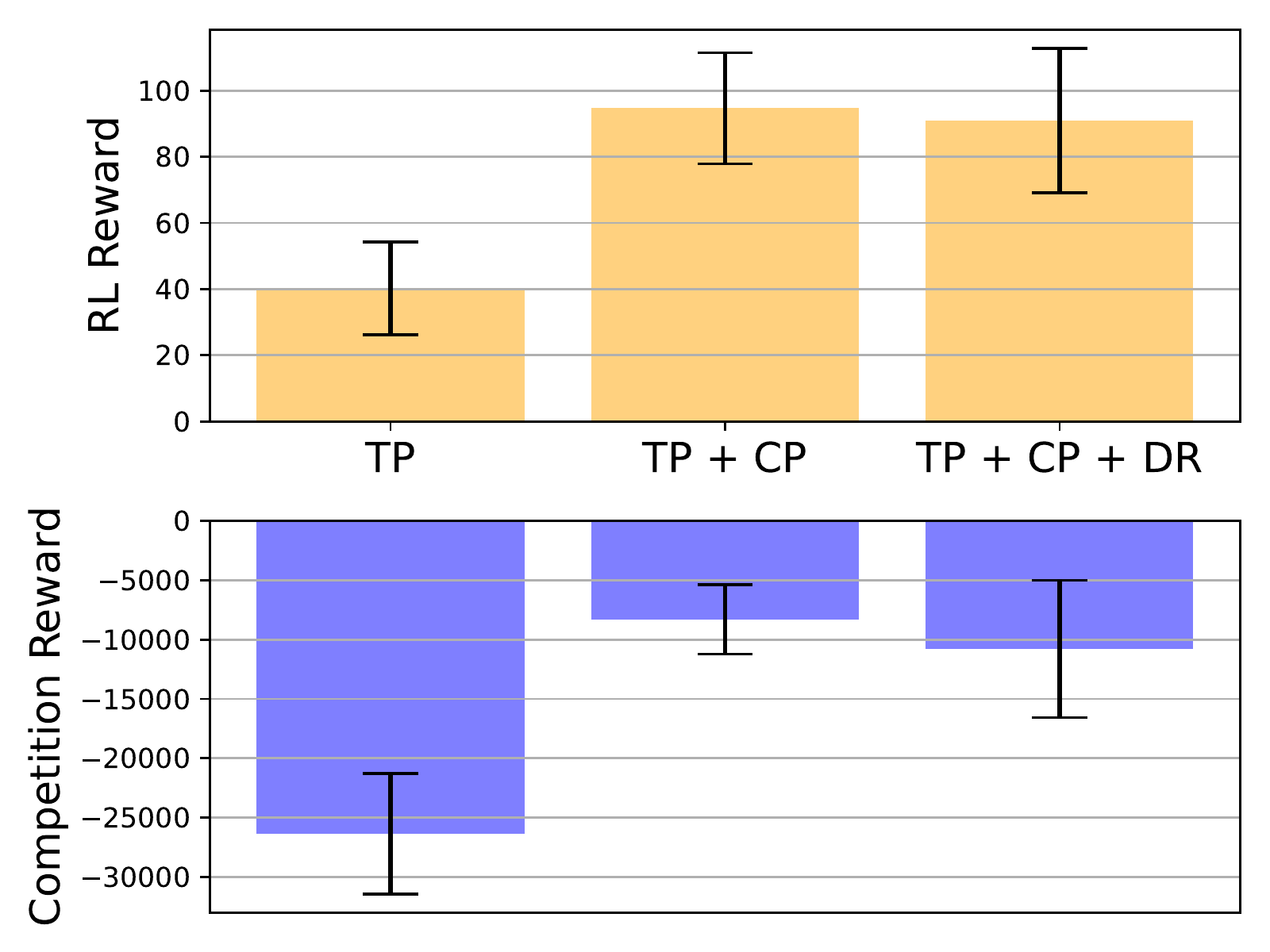}
\caption{Rewards comparison of different policies. From left to right: Trajectory Planning(TP) only, TP + Contact Planning(CP), TP+CP+Domain Randomization(DR). Top bar-graph shows the reward for RL training, and bottom one shows the reward of the contest}
\label{fig:result}
\end{figure}

We test the performance of our method by training in a simulation environment and running multiple experiments in different situations in the simulation and real systems. In the results of the experiments, our method is compatible with both stability and adaptability as shown in Fig. \ref{fig:result}. 

At the same time, due to the addition of fifth-order polynomials, the movement of the square is relatively smooth and no jitters. At the same time, high-level reinforcement learning can select contact points suitable for the current situation according to different positions and states, which enables the model to have stronger adaptability. Moreover, this hierarchical control method reduces the need for training samples and training time in the training process and alleviates the sim-real problems. Our method demonstrates the potential of reinforcement learning and control methods in robotic tasks to a certain extent.

\begin{figure}[ht]
\centering

\includegraphics[width=0.25\textwidth]{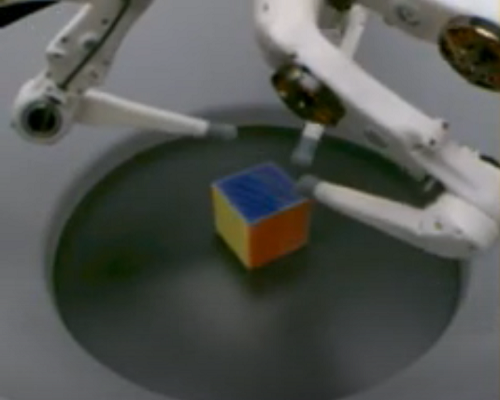}
\includegraphics[width=0.25\textwidth]{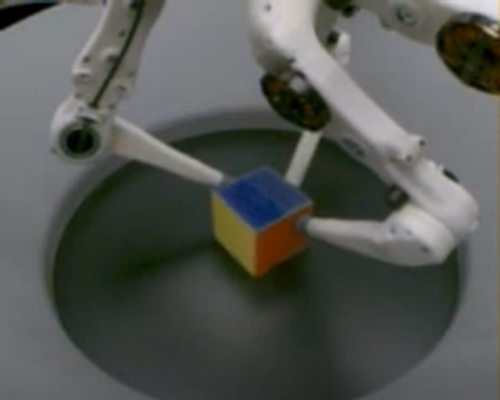}
\includegraphics[width=0.25\textwidth]{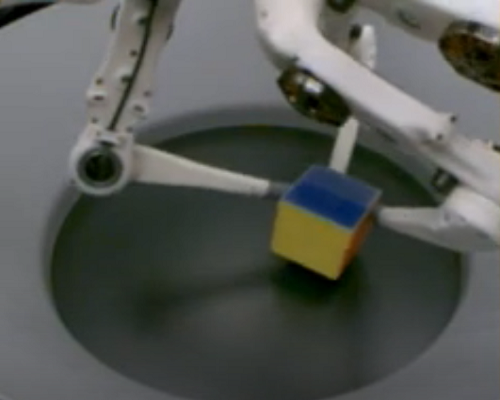}
\caption{Physical experiment}
\label{Fig.real}
\end{figure}

We directly deploy the model trained in the simulation to the real environment, and the experiment proves that our method can be applied to the real robot. This is because we combine reinforcement learning with model-based controller, which has better stability as shown in \ref{Fig.real}.  

The next step of the research needs to consider the challenges that the algorithm faces small. The algorithm needs to maintain the awareness of environmental changes, such as how to avoid collisions with the dice that have been arranged and complete high-level planning. We plan to introduce world models\cite{ha2018world,hafner2019dream,hafner2020mastering} to increase the ability of the agent to control the environment. When there is a big difference between simulation and reality, we will also add the sim-real algorithm\cite{peng2018sim,andrychowicz2020learning,chebotar2019closing,james2019sim}.

\bibliographystyle{unsrt}
\bibliography{references}  

\begin{thebibliography}{10}

\bibitem{wuthrich2020trifinger}
Manuel W{\"u}thrich, Felix Widmaier, Felix Grimminger, Joel Akpo, Shruti Joshi,
  Vaibhav Agrawal, Bilal Hammoud, Majid Khadiv, Miroslav Bogdanovic, Vincent
  Berenz, et~al.
\newblock Trifinger: An open-source robot for learning dexterity.
\newblock {\em arXiv preprint arXiv:2008.03596}, 2020.

\bibitem{haarnoja2018soft}
Tuomas Haarnoja, Aurick Zhou, Pieter Abbeel, and Sergey Levine.
\newblock Soft actor-critic: Off-policy maximum entropy deep reinforcement
  learning with a stochastic actor.
\newblock In {\em International Conference on Machine Learning}, pages
  1861--1870. PMLR, 2018.

\bibitem{lynch2017modern}
Kevin~M Lynch and Frank~C Park.
\newblock {\em Modern robotics}.
\newblock Cambridge University Press, 2017.

\bibitem{ha2018world}
David Ha and J{\"u}rgen Schmidhuber.
\newblock World models.
\newblock {\em arXiv preprint arXiv:1803.10122}, 2018.

\bibitem{hafner2019dream}
Danijar Hafner, Timothy Lillicrap, Jimmy Ba, and Mohammad Norouzi.
\newblock Dream to control: Learning behaviors by latent imagination.
\newblock {\em arXiv preprint arXiv:1912.01603}, 2019.

\bibitem{hafner2020mastering}
Danijar Hafner, Timothy Lillicrap, Mohammad Norouzi, and Jimmy Ba.
\newblock Mastering atari with discrete world models.
\newblock {\em arXiv preprint arXiv:2010.02193}, 2020.

\bibitem{peng2018sim}
Xue~Bin Peng, Marcin Andrychowicz, Wojciech Zaremba, and Pieter Abbeel.
\newblock Sim-to-real transfer of robotic control with dynamics randomization.
\newblock In {\em 2018 IEEE international conference on robotics and automation
  (ICRA)}, pages 3803--3810. IEEE, 2018.

\bibitem{andrychowicz2020learning}
OpenAI:~Marcin Andrychowicz, Bowen Baker, Maciek Chociej, Rafal Jozefowicz, Bob
  McGrew, Jakub Pachocki, Arthur Petron, Matthias Plappert, Glenn Powell, Alex
  Ray, et~al.
\newblock Learning dexterous in-hand manipulation.
\newblock {\em The International Journal of Robotics Research}, 39(1):3--20,
  2020.

\bibitem{chebotar2019closing}
Yevgen Chebotar, Ankur Handa, Viktor Makoviychuk, Miles Macklin, Jan Issac,
  Nathan Ratliff, and Dieter Fox.
\newblock Closing the sim-to-real loop: Adapting simulation randomization with
  real world experience.
\newblock In {\em 2019 International Conference on Robotics and Automation
  (ICRA)}, pages 8973--8979. IEEE, 2019.

\bibitem{james2019sim}
Stephen James, Paul Wohlhart, Mrinal Kalakrishnan, Dmitry Kalashnikov, Alex
  Irpan, Julian Ibarz, Sergey Levine, Raia Hadsell, and Konstantinos Bousmalis.
\newblock Sim-to-real via sim-to-sim: Data-efficient robotic grasping via
  randomized-to-canonical adaptation networks.
\newblock In {\em Proceedings of the IEEE/CVF Conference on Computer Vision and
  Pattern Recognition}, pages 12627--12637, 2019.

\end{thebibliography}






\end{document}